%% file: icra16mastalli.tex
\documentclass[letterpaper, 10 pt, conference]{ieeeconf}
\IEEEoverridecommandlockouts
\usepackage{graphics}
\usepackage[tight]{subfigure}
\usepackage[pdftex]{graphicx}
\usepackage{amsfonts}
\usepackage{amsmath}
\usepackage{url}
\usepackage{epstopdf}
\usepackage{multirow}
\usepackage[pdfpagemode=UseNone,pdftoolbar=false,pdfauthor=Carlos Mastalli /
Ioannis Havoutis / Michelle Focchi,pdfstartview=FitH]{hyperref}
\hypersetup{urlcolor=blue, colorlinks=true, linkcolor=black}
\usepackage{color}
\usepackage[usenames,dvipsnames]{xcolor}
\usepackage{bm}
\usepackage[normalem]{ulem}
\usepackage[leftcaption]{sidecap}
\usepackage{booktabs}
\usepackage{vector}
\usepackage{glossaries}
	\newacronym{hyq}{HyQ}{Hydraulically Actuated Quadruped}
	\newacronym{cog}{CoG}{Center of Gravity}
	\newacronym{com}{CoM}{Center of Mass}
	\newacronym{cmm}{CMM}{Centroidal Momentum Matrix}
	\newacronym{mpcc}{MPCC}{Mathematical Program with Complementarity Constraints}
	\newacronym{sqp}{SQP}{Sequential Quadratic Programming}
	\newacronym{zmp}{ZMP}{Zero Moment Point}
	\newacronym{dof}{DoF}{Degree of Freedom}
\usepackage[tight]{units}

\newcommand\eat[1]{}

\newcommand{\sref}[1]{Section~\ref{#1}}
\newcommand{\fref}[1]{Fig.~\ref{#1}}

\newcommand{\tref}[1]{Table~\ref{#1}}

\renewcommand{\baselinestretch}{0.98}%

\input{math_commands.tex}
\begin{document}
\title{\LARGE \bf
Hierarchical Planning of Dynamic Movements \\ without Scheduled Contact
Sequences}

\author{{\centering
Carlos Mastalli$^1$, \quad Ioannis Havoutis$^2$, \quad Michele
Focchi$^1$,}\\\quad Darwin G. Caldwell$^1$, \quad Claudio Semini$^1$%
\thanks{
\hspace{-1em}$^1$Department of Advanced Robotics, Istituto Italiano di
Tecnologia, via Morego, 30, 16163 Genova, Italy. \textit{email}: \{carlos.mastalli,
michele.focchi, darwin.caldwell, claudio.semini\}@iit.it.
\newline
$^2$Robot Learning and Interaction Group, Idiap Research Institute, Martigny,
Switzerland \textit{email}: ioannis.havoutis@idiap.ch
\newline
This work was in part supported by the DexROV project through the EC
Horizon 2020 programme (Grant \#635491).
}
}

\maketitle
\thispagestyle{empty}
\pagestyle{empty}

\begin{abstract}
Most animal and human locomotion behaviors for solving complex tasks involve
dynamic motions and rich contact interaction. In fact, complex maneuvers need to
consider dynamic movement and contact events at the same time. We present a
hierarchical trajectory optimization approach for planning dynamic movements
with unscheduled contact sequences. We compute whole-body motions that achieve
goals that cannot be reached in a kinematic fashion. First, we find a feasible
CoM motion according to the centroidal dynamics of the robot. Then, we refine
the solution by applying the robot's full-dynamics model, where the feasible CoM
trajectory is used as a warm-start point. To accomplish the unscheduled contact
behavior, we use complementarity constraints to describe the contact model, i.e.
environment geometry and non-sliding active contacts. Both optimization phases
are posed as \gls{mpcc}. Experimental trials demonstrate the performance of our
planning approach in a set of challenging tasks.

\end{abstract}

\input{src/introduction}

\input{src/related_work}

\input{src/planning}

\input{src/experimental_results}

\input{src/conclusion}

\bibliographystyle{./IEEEtran}
\bibliography{ref/planning,ref/control,ref/learning,ref/locomotion,ref/optimization,ref/perception,ref/software,ref/robot}

\end{document}

%% file: math_commands.tex
\newcommand{\mx}[1]{\mathbf{\bm{#1}}} 				%
\newcommand{\vc}[1]{\mathbf{\bm{#1}}} 					%
\newcommand{\mat}[1]{\ensuremath{\begin{bmatrix}#1\end{bmatrix}}}	%

%% file: src/introduction.tex
\section{Introduction}\label{sec:introduction}
Legged robots are able to traverse areas that are inaccessible to wheeled
vehicles, such as complex and unstructured environments. Such environments are
common to search and rescue scenarios, one promising application of legged
systems. From the legged locomotion point of view, natural disaster scenarios
require planning and execution of complex behaviors in environments with high
uncertainty. Complex behaviors require whole-body movements and multiple
contact interactions at the same time. Indeed, whenever an articulated body is
in contact with the environment (e.g. legged robots) the set of contact forces
and joint commands describe the evolution of the motion. Dynamic maneuvers such
as jumping and rearing need to consider different sequences of contacts
(mode-switching), which cannot be tackled with traditional predefined gaits \cite{Barasuol2013,Winkler2014}.

Locomotion in complex environments requires reasoning about terrain conditions,
planning and execution of movements through a sequence of contacts, i.e.
footholds and/or handholds. These can be posed as separate problems (decoupled
approach), i.e. motion and contact planning, and control. Such approaches reduce
the combinatorial search space at the expense of the richness of complex
behaviors. In contrast, highly-dynamic movements need to consider contact forces
and robot dynamics. For instance, contact forces play an important role for
predicting the ballistic trajectory in a jumping task. A decoupled motion
planner has to explore different plans in the space of feasible movements, which
is often defined by physical (friction properties), stability (whole-body
balance), dynamic (inertial properties) and task constraints (goal positions and
orientations). On the other hand, coupled motion planners compute simultaneously
contacts and body movements by posing the problem as a hybrid system or a
mode-invariant optimization problem.

\begin{figure}[t]
	\centering
	\includegraphics[width=0.9\columnwidth]{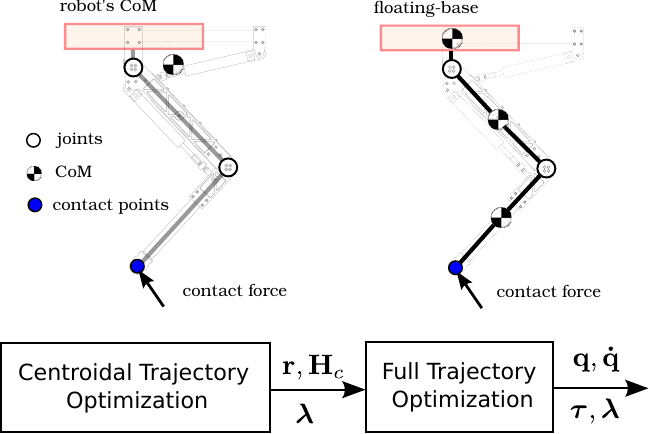}
	\caption{The proposed hierarchical trajectory optimization reduces the
	complexity of the motion planning problem by considering two different
	optimization
	phases: centroidal and full trajectory optimization. First, the centroidal
	trajectory optimization phase produces a locally optimal \gls{com}
	motion using the centroidal dynamics model \cite{Orin2013}, which does not
	consider joint dynamics (i.e. link's \gls{com}). Second, the full trajectory
	optimization phase refines the \gls{com} trajectory by applying the
	robot's full-dynamics and joint limits. Both optimization phases
	use complementarity constraints to model the contact interactions.}
\label{fig:hierarchical_trajectory_opt}
\end{figure}

In this paper, we are concerned with finding feasible trajectories for complex
tasks, i.e. tasks that require the exploration of different mode sequences
through highly-dynamic movements. We choose a set of jumping tasks as examples,
as these highlight the ability to explore the dynamical capabilities of 
the robot in order to reach goals that are unreachable in a kinematic manner.

The main contribution of this paper is a novel hierarchical trajectory
optimization approach based on the principle of \textit{divide and conquer} (
Fig. \ref{fig:hierarchical_trajectory_opt}). Our hierarchical trajectory
optimization is capable of producing a wide range of complex behaviors without
scheduling a contact sequence by reducing the search space to a fixed sequence
of commands, which can be used to find continuous motion plans. The trajectory
optimization approach is posed as a \gls{mpcc}. First, we prune the search space
by finding a feasible trajectory according to the robot's centroidal dynamics.
Second, we find a feasible trajectory in terms of the full dynamics and joint
limits of the robot.

The rest of the paper is structured as follows: after discussing previous
research in the field of dynamic motion planning for legged systems
(\sref{sec:related_work}) we describe the direct hierarchical trajectory
optimization approach proposed for dynamic motion planning in
\sref{sec:planning}. In \sref{sec:experimental_results} we evaluate the
performance of our trajectory optimization approach in experimental trials with
a robotic leg before \sref{sec:conclusion} summarizes this work and presents
ideas for future work.

%% file: src/related_work.tex
\section{Related Work}\label{sec:related_work}
State-of-the-art legged motion planning methods are often constrained to
non-dynamic regimes with scheduled contact sequences
\cite{Kolter2008b}\cite{Kalakrishnan2010a}\cite{Havoutisi2013}\cite{Winkler2015} and decoupled
models \cite{Mastalli2015}\cite{Dellin2012}. For example, most of the current
locomotion approaches follow predefined contact sequences, specified by fixed
patterns of locomotion. These assumptions reduce the capability of traversing
complex environments with legged systems. We believe that natural locomotion in
complex environments requires dynamic movements with unscheduled contact
sequences. Nevertheless, these approaches are often posed as non-linear
optimization problems in which kinematic and dynamic models of the robot, and
task information are stated as hard
\cite{Posa2013}\cite{Dai2014}\cite{Yuval2010} or soft constraints
\cite{Mordatch2012a}\cite{Gabiccini2015}. Currently, there are many approaches
that reduce the complexity of the problem. Different dynamic formulations have
been proposed such as: point-mass, cart-table
\cite{Kalakrishnan2010a}\cite{Winkler2015}, spring-loaded inverted pendulum
model \cite{Poulakakis2009}\cite{Erez2013} and contact wrench sum
\cite{Dai2014}\cite{Orin2013}. However, most of these approaches use simplified
models that do not capture the full-body dynamics of the robot, and reduce the
richness of the solutions. On the other hand, using full-body dynamic models
\cite{Posa2013}\cite{Tassa2012}, one can produce richer movements but often such
formulations require an excessively long amount of time to compute solutions,
they are prone to reaching and becoming trapped in local minima and
infeasible regions.

Defining a scheduled contact sequence considerably reduces the search space by
constraining the plans to a fixed pattern of locomotion \cite{Mastalli2015}.
On the other hand, with an unscheduled contact sequence, we can increase the
possibilities of locomotion \cite{Hauser2011}\cite{Escande2013}, which could in
turn be crucial for finding solutions in complex environments. From the
optimization point of view, unscheduled contact sequence approaches include the
contact forces as decision variables, allowing us to consider friction cone
constraints. In the literature there exist two main approaches that incorporate
contact forces in the optimization; the first one, smooths the contact forces
but has the drawback of allowing the contact to be enabled when there is a
distance w.r.t. the contact surface \cite{Mordatch2012a}, the second one, uses
complementarity constraints to model inelastic contacts
\cite{Posa2013}\cite{Dai2014}. Complementarity constraints describe inelastic
contact forces without resulting in stiff differential equations that require an
extremely small time step \cite{Stewart2000}.

%% file: src/planning.tex
\section{Hierarchical Dynamic Planning}\label{sec:planning}
Our work was motivated by the observation that most animal and human locomotion
behaviors involve dynamic motions through contact interactions. For instance,
kangaroos are \textit{dynamic jumpers} that use hopping as the main locomotion
strategy. Indeed, in kangaroo locomotion, highly-dynamic movements and contact
forces play an important role for finding efficient locomotion trajectories.

Although such dynamic maneuvers are undoubtedly beneficial, planning and
execution of these whole-body trajectories is challenging due to the loss of
control authority during flight phases and undefined contact events. We tackle it
by generating a whole-body trajectory toward a body goal state (desired body
height) that ensures a dynamic motion plan through contact interactions. To
accomplish this, we describe the contact model using complementarity
constraints which defines our approach as a mode-invariant trajectory
planner.

\subsection{Generating Dynamic Motions}
Consider a rigid body system with $n$ degrees of freedom, of which $n_b$ are
floating-base degrees of freedom. The state of the robot is represented by its
floating-base and actuated joint components, $\vc{q}=(\vc{q}_b,\vc{q}_q)$.
Additionally, the robot has $p$ end-effector or contact points.

The system's evolution depends on the internal joint torques, $\vc{\tau}_q$, and
the contact force, $\vc{\lambda}_j$, applied at the $j^{th}$ end-effector. This
evolution is subject to robot and environmental constraints such as: joint
limits and environment geometry. Exploring different mode switches (contact
events) and dynamic movements could produce unsuccessful locally optimal
solutions. We improve the solutions by applying a hierarchical trajectory
optimization. In the first phase, we model the system's evolution with the
centroidal dynamics, i.e. in the \gls{com} space. Then, we impose joint dynamics
and limits using a full-dynamic model. Figure
\ref{fig:hierarchical_trajectory_opt} presents an overview of our hierarchical
trajectory optimization approach.

\subsubsection{Centroidal-dynamic model}
The robot dynamics can be projected at the \gls{com}, i.e. the
\textit{centroidal dynamics} of the robot. In a full floating-base system ($n_b
= 6$ DoF), the centroidal-dynamic model describes the rate of change of linear
and angular momentum of \gls{com} with respect to the inertial frame of
reference \cite{Orin2013}. The rate of change of linear and angular momentum is
determined by contact forces $\vc{\lambda}_j$, gravitational force $m\vc{g}$ and
the motion of the robot's links
\begin{eqnarray}
	m\vc{\ddot{r}} = \sum_{j=0}^p \vc{\lambda}_j + m\vc{g} \\
	\mx{\dot{H}}_c(\vc{q},\vc{\dot{q}}) = \sum_{j=0}^p (\vc{x}_j - \vc{r}) \times
	\vc{\lambda}_j
\label{eq:centroidal_dynamics}
\end{eqnarray}
where $m$ is the total mass of the robot, $\vc{r}\in\mathbb{R}^3$ is the
\gls{com} position, $\vc{\lambda}_j\in\mathbb{R}^3$ is the contact force applied
at the $j^{th}$ end-effector, $\mx{H}_c\in\mathbb{R}^3$ is the centroidal
angular momentum and $\vc{x}_j\in\mathbb{R}^3$ is the end-effector position. The
centroidal angular momentum is computed through the computation of the \gls{cmm}
as defined in \cite{Orin2013}.

\subsubsection{Full-dynamic model}
The full-dynamic model enables us to compute the joint efforts given a motion
state $(\vc{q},\vc{\dot{q}},\vc{\ddot{q}})$ subject to contact forces
$\vc{\lambda}_j$. We partition the dynamics equation of the robot into the
unactuated floating-base DoFs $\vc{q}_b$ ($n_b$ equations) and the active robot
joints $\vc{q}_q$ ($n_q$ equations):
\begin{equation}
	\underbrace{\mx{H}(\vc{q}) \mat{\vc{\ddot{q}}_b \\ \vc{\ddot{q}}_q} + 
	\mat{\vc{c}_b \\ \vc{c}_q}(\vc{q}, \vc{\dot{q}}) - \sum_{j=0}^p
	\mat{\mx{J}_{b_j}^T
	\\
	\mx{J}_{q_j}^T}\vc{\lambda}_j}_{\vc{b} =
	ID(model,\vc{q},\vc{\dot{q}},\vc{\ddot{q}})} = \mat{\vc{0}\\
	\vc{\tau}_q}
\label{eq:full_dynamics}
\end{equation}
where $\mx{H}\in\mathbb{R}^{n\times n}$ is the floating-base inertial matrix,
$\vc{c} = (\vc{c}_{b}, \vc{c}_{q})\in\mathbb{R}^n$ is the force vector that
accounts for Coriolis, centrifugal, and gravitational forces,
$\vc{\lambda}_j\in\mathbb{R}^3$ are the ground contact forces at the $j^{th}$
end-effector (i.e. point feet), and their corresponding Jacobian, $\mx{J}_j =
\mat{\mx{J}_{b_j} & \mx{J}_{q_j}}\in\mathbb{R}^{3p\times n}$ and
$\vc{\tau}_q\in\mathbb{R}^{n_q}$ are the joint efforts that we wish to
calculate.

The left-hand term $\vc{b} = (\vc{b}_{b}, \vc{b}_{q})$ is computed efficiently
using the Featherstone implementation of the Recursive Newton-Euler Algorithm
(RNEA) \cite{Featherstone2008}.

\subsection{Contact Model}
In dynamic movements, contact forces play an important role, e.g. a jumping or
hopping task. Traditional approaches compute trajectories given a predefined
contact sequence. These approaches do not exploit the fact that an optimized
mode switching could be required for the success of a certain task.

A contact event occurs when a signed distance to the surface is strictly zero,
and additionally, there is a contact force acting along the surface normal.
Moreover, it is expected a null normal contact force when the contact is
inactive, i.e. a positive signed distance. In other words, normal contact forces
and signed distances are orthogonal and positives functions
(\ref{eq:contact_model}). Additionally, we desire that active contacts do not
slide. Such condition implicates an orthogonality between normal contact forces
and tangential velocities (\ref{eq:contact_velocity}). In the optimization
literature, constraints with combinatorial nature, such as the set of contact
model equations (\ref{eq:contact_model})(\ref{eq:contact_velocity}), can be
described as complementarity constraints
\begin{eqnarray}
	&0 \leq \lambda_j^{\hat{n}} \perp \phi_j(\vc{q}) \geq 0
	\label{eq:contact_model}\\\
	&\vc{0} \leq \lambda_j^{\hat{n}} \perp
	\vc{\dot{x}}_j^{\hat{t}}(\vc{q},\vc{\dot{q}}) \geq \vc{0}
	\label{eq:contact_velocity}
\end{eqnarray}
where $\lambda_j^{\hat{n}}$ is the contact force acting along the surface normal
at the $j^{th}$ end-effector (i.e. contact point), $\phi_j(\mathbf{q})$ is the
signed distance between the $j^{th}$ contact point $\vc{x}_j$ and the surface
$\mathcal{S}_i$, and $\vc{\dot{x}}_j^{\hat{t}}(\vc{q},\vc{\dot{q}})$ is the
velocity of the contact point along the tangential surface. Contact-point
positions and velocities are calculated efficiently using spatial algebra.

\subsection{Trajectory Optimization}
Planning problems without scheduled contact sequences are often hard to solve
due to the contact forces producing discontinuities in the dynamics. Here, we
tackle this issue by applying a hierarchical trajectory optimization scheme,
which uses different dynamic models in a two-phase manner. Using a different
(simpler) dynamic model in the first optimization phase, we imposes a dynamic
relaxation, that helps to explore different mode switches. Thus, we find a
feasible \gls{com} motion in terms of the robot's centroidal dynamics. Then, we
refine it by applying the full-dynamic model in the second, more complex,
trajectory optimization phase.

\subsubsection{Centroidal trajectory optimization}
The centroidal trajectory optimization step computes a feasible \gls{com} trajectory
through the mapping of contact forces inside the centroidal dynamics. The
\gls{cmm} maps the robot's generalized velocities to its spatial momentum (for
more details see \cite{Orin2013}). We sample the trajectory in $N$ knot-points
with a fixed-time duration $h$. In this optimization phase, the decision
variables of the optimization problem are the robot position $\vc{q}$, the robot
velocity $\vc{\dot{q}}$, the \gls{com} position $\vc{r}$, the \gls{com} velocity
$\vc{\dot{r}}$, the contact forces $\vc{\lambda}$, and the end-effector
(contact) positions $\vc{x}$. Our cost function evaluates the trajectory in
terms of the desired high-level goal of the task $\vc{w}(\vc{q})$ as:
\begin{equation}
	\min_{\substack{\vc{q}[k],\vc{\dot{q}}[k],\vc{r}[k],\vc{\dot{r}}[k],\\
	\vc{H}_c[k],\vc{\dot{H}}_c[k],\vc{\lambda}[k],\vc{x}[k]}} h\sum_{k=1}^N
	\left(\|\vc{w}(\vc{q}[k]) - \vc{w}(\vc{q}^*[k])\|_{\mx{Q}_q}\right)
\end{equation}
where $\vc{w}(\vc{q})$ constructs a task-specific value from relevant features
of the task, and $\|\vc{w}(\vc{q}[k]) - \vc{w}(\vc{q}^*[k])\|_{\mx{Q}_q}$
computes its associated quadratic cost given a desired robot position
$\vc{q}^*$. Note that $\|\mathbf{x}\|_\mx{Q}$ is an abbreviation for the
quadratic cost $\mathbf{x}^T\mx{Q}\mathbf{x}$.

We transcribe the centroidal dynamics differential equations
(\ref{eq:centroidal_dynamics}) to algebraic ones by applying an Euler-backward
integration rule with a fixed-time step $h$
\begin{eqnarray}
	\vc{r}[k-1] - \vc{r}[k] + h\vc{\dot{r}}[k] = \vc{0} \\
	\vc{H}_c[k-1] - \vc{H}_c[k] + h\vc{\dot{H}}_C[k] = \vc{0} \\
	m(\vc{\dot{r}}[k]-\vc{\dot{r}}[k-1]) - h\Bigg(\sum_{j=0}^p \vc{\lambda}_j[k] +
	m\vc{g}\Bigg) = \vc{0} \\
	\mx{\dot{H}}_c[k] - \sum_{j=0}^p (\vc{x}_j[k]-\vc{r}[k])\times
	\vc{\lambda}_j[k] =
	\vc{0}
\label{eq:centroidal_dynamics_transcription}
\end{eqnarray}
where the centroidal angular momentum is computed from the \gls{cmm},
$\mx{A}(\vc{q})$, as is explained in \cite{Orin2013}, i.e. $\mx{H}_c[k]=
\mx{A}(\vc{q}[k])\vc{\dot{x}}[k]$. Additionally, we impose contact position
constraints in order to describe the contact interactions
\begin{equation}
	\vc{x}_j[k] - \vc{\kappa}_j(\vc{q}[k]) = \vc{0}
\label{eq:contact_position_constraints} 
\end{equation}
where $\vc{\kappa}_j(\cdot)$ is the direct kinematic function which computes the
position of the $j^{th}$ end-effector. We also include joint position and
velocity limits.
\begin{eqnarray}
	\vc{q}_q^l \leq \vc{q}_q \leq \vc{q}_q^u
\label{eq:centroidal_bounds_1}\\
	\vc{\dot{q}}_q^l \leq \vc{\dot{q}}_q \leq \vc{\dot{q}}_q^u
\label{eq:centroidal_bounds_2}
\end{eqnarray}

To describe different possible mode switches, we add contact position and
velocity constraints. These constraints are described as complementarity
constraints as follows
\begin{eqnarray}
	\lambda_j^{\hat{n}}[k], \phi_j(\vc{q}[k]) \geq 0
\label{eq:complement_functions} \\ 
	\lambda_j^{\hat{n}}[k] \phi_j(\vc{q}[k]) = 0
\label{eq:position_complementarity_constraints} \\
	\lambda_j^{\hat{n}}[k] \left(\vc{x}_j^{\hat{t}}[k] -
	\vc{x}_j^{\hat{t}}[k-1]\right) = \vc{0}
\label{eq:velocity_complementarity_constraints}
\end{eqnarray}

We approximate the contact velocity as contact displacement along the
tangential surface. Note that a contact velocity constraint does not guarantee
zero displacement between knots.

\subsubsection{Full trajectory optimization}
Once a feasible, and locally optimal, \gls{com} trajectory is computed, we use
this \gls{com} trajectory as a warm-start point of the full trajectory
optimization phase. We transcribe the full-dynamic model with the same time step
value of the centroidal trajectory optimization phase. In this optimization
phase, we formulate the problem with the following decision variables: the robot
position $\vc{q}$, the robot velocity $\vc{\dot{q}}$, the joint efforts
$\vc{\tau}_q$ and the contact forces $\vc{\lambda}$.

In this stage, the cost function also considers the joint effort energy of the
movement $\vc{\tau}_q$ as
\begin{equation}
	\min_{\substack{\vc{q}[k],\vc{\dot{q}}[k],\\ \vc{\tau}[k],\vc{\lambda}[k]}}
	h\sum_{k=1}^N \left(\|\vc{w}(\vc{q}[k]) - \vc{w}(\vc{q}^*[k])\|_{\mx{Q}_q} +
	\|\vc{\tau}_q[k]\|_{\mx{R}}\right)
\label{eq:full_optimization_cost}
\end{equation}

We apply the same integration rule to the full-dynamic differential equation
(\ref{eq:full_dynamics}). Additionally, we add a selection matrix $\mx{B}$ in
order to impose a null wrench vector to the floating-base:
\begin{equation}
	\vc{q}[k-1] - \vc{q}[k] + h\vc{\dot{q}}[k] = \vc{0}
\end{equation}
\begin{equation}
\begin{split}
	\mx{H}[k] \left(\vc{\dot{q}}[k] - \vc{\dot{q}}[k-1]\right) \\
	 + h\Big(\vc{c}[k] - \sum_{j=0}^p \mx{J}_j&[k]^T \vc{\lambda}_j[k]\Big)
	 - \mx{B}\vc{\tau}[k] = \vc{0}
\label{eq:full_dynamics_transcription}
\end{split}
\end{equation}
where the contact forces are determined using the complementarity constraints
(\ref{eq:complement_functions})(\ref{eq:position_complementarity_constraints})(\ref{eq:velocity_complementarity_constraints}).

In the full trajectory optimization phase, we impose position and velocity
bounds (\ref{eq:centroidal_bounds_1})(\ref{eq:centroidal_bounds_2}), and
additionally joint efforts bounds
\begin{eqnarray}
	\vc{\tau}_q^l \leq \vc{\tau}_q \leq \vc{\tau}_q^u
\label{eq:full_bounds}
\end{eqnarray}

We derive a continuous motion plan, from the $N$ optimized knot-points, using a
polynomial interpolation. Both optimization phases model contact interactions
using complementarity constraints. In general, optimization problems with
complementarity constraints are difficult to solve because constraint
qualifications are hard to satisfy. We solve the \gls{mpcc} using interior point
method as this is faster than a \gls{sqp} algorithm when the number of
complementarity constraints increases \cite{Raghunathan2005}. We use the
\textsc{Ipopt} library \cite{Wachter2006}. We relax the orthogonality between
the complementarities, for example $\lambda_j^{\hat{n}}[k] \phi_j(\vc{q}[k]) =
0$ is posed as $\lambda_j^{\hat{n}}[k] \phi_j(\vc{q}[k]) \leq 0$. For more
information about different interior point methods see \cite{Raghunathan2005}.

%% file: src/experimental_results.tex
\section{Experimental Results}\label{sec:experimental_results}
This section describes the experiments conducted to validate the effectiveness 
and quantify the performance of the proposed hierarchical optimization approach.

\subsection{Experimental Setup}
\begin{figure}%
	\centering
	\subfigure{\resizebox{0.425\columnwidth}{!}{
	\includegraphics{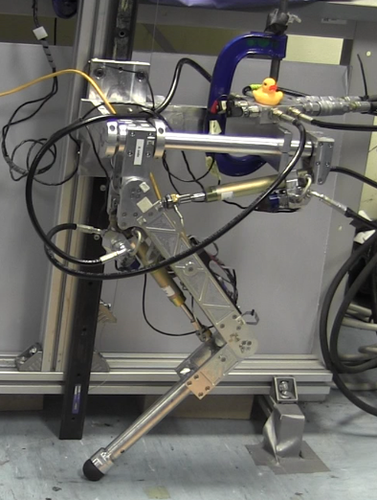}}}
	\subfigure{\resizebox{0.425\columnwidth}{!}{
	\includegraphics{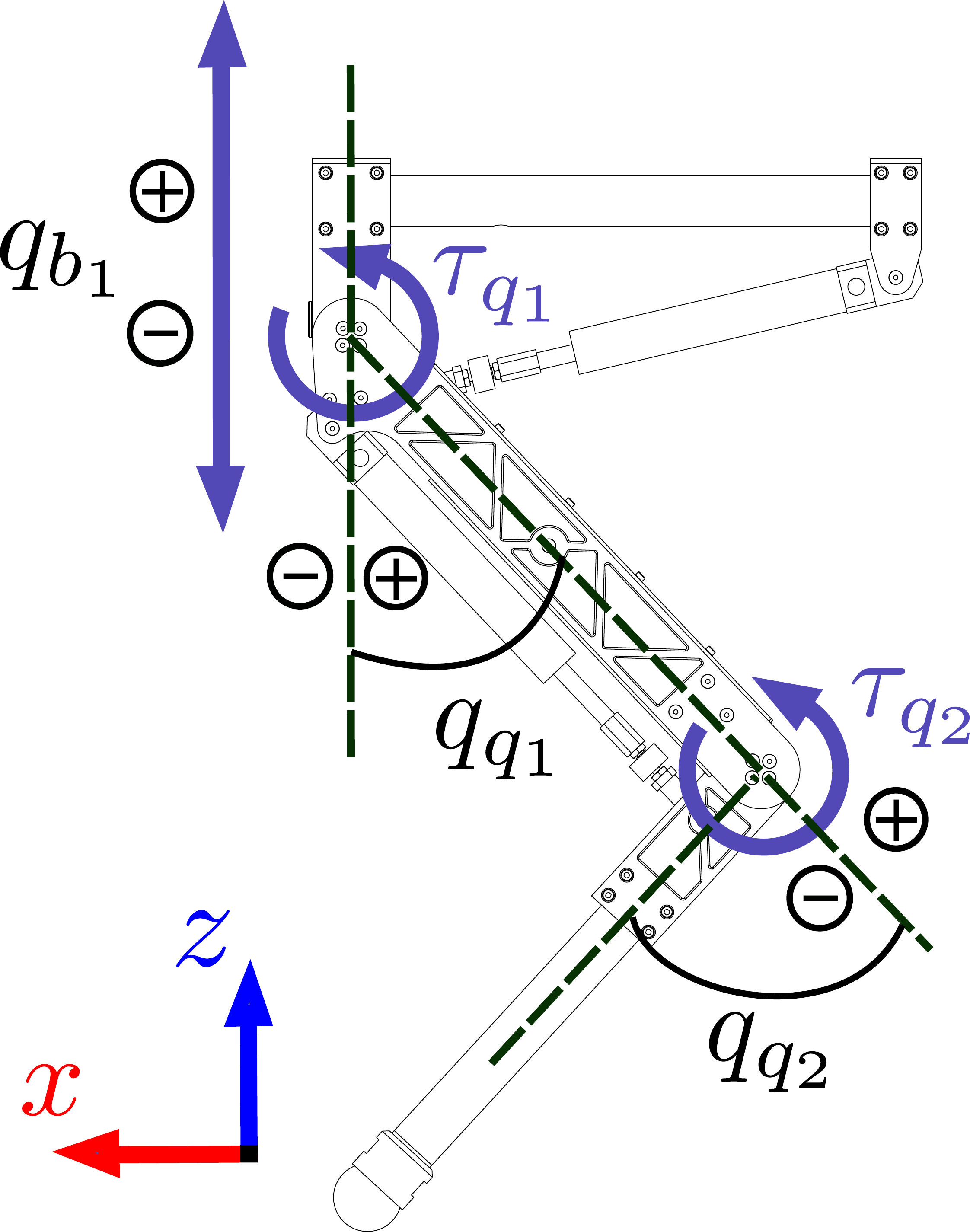}}}
	\caption{HyL: one hydraulically actuated and fully torque controlled leg of the
	quadruped robot HyQ \cite{Semini2011}. The HyL robot has a total number of 3
	DOF: $1$D floating-base system $q_{b_1}$ with 2 actuated joints
	$(q_{q_1},q_{q_2})$.}\vspace{-0.2cm}
\label{fig:hyl}
\end{figure}

We use the hydraulically-actuated robot leg, HyL, in our experiments. HyL weighs approximately 11 kg, is fully-torque
controlled and equipped with precision joint encoders, and load cells. HyL is a
$1$D floating-base system with 2 actuated joints as is shown in Fig.
\ref{fig:hyl}. Controller computations are done on-board in an i7/2.8 GHz PC 
with a realtime-time kernel. Motion plans are computed off-line using a
predefined terrain model.

\begin{figure*}[htb]
	\centering
	\subfigure{\resizebox{0.96\textwidth}{!}{
	\includegraphics{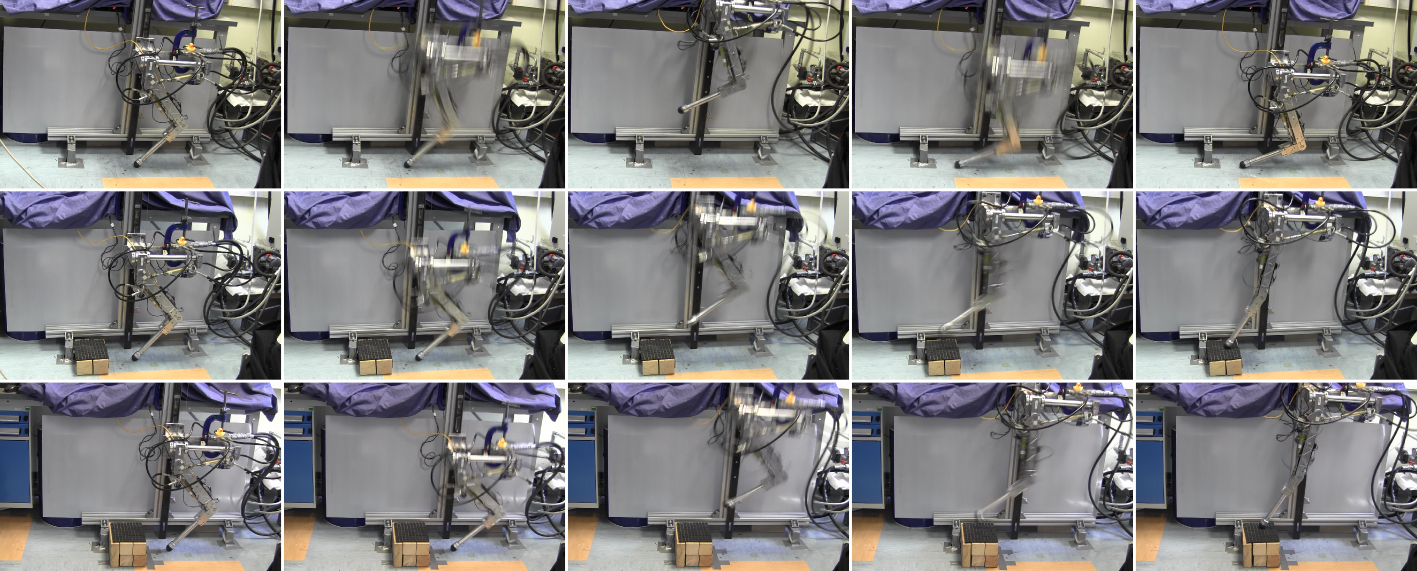}}}
	\caption{Snapshots of three experimental trials used to evaluate the
	performance of our hierarchical trajectory optimization approach. From top to
	bottom:	jumping task; small step jumping (\unit[10]{cm} of height); big step
	jumping (\unit[15]{cm} of height).}
	\label{fig:vidsnaps}
\end{figure*}

For each experiment, we specify the goal state of the robot's trunk, and the
desired final joint position as a terminal cost. The hierarchical trajectory
optimizer finds a sequence of footsteps through dynamic movements without a
predefined order, which the controller then executes dynamically. We use a PD
controller, using the planned joint efforts as feedforward inputs. 
We validate the
performance of our framework in 3 different examples as seen in
\fref{fig:vidsnaps}, and compare against the full dynamic optimization
(\tref{tab:results}) on the same benchmark examples. The first example consists
of reaching a goal that is kinematically not feasible, called the \textit{jumping
task}. In the next examples, the \textit{step-jumping tasks}, the robot has 
to reach
and keep the desired trunk height, which is done through two different steps:
a small step (\unit[10]{cm}) and a big step (\unit[15]{cm}). 
Additionally, the reader is
strongly encouraged to view the accompanying video
as it provides the most intuitive way to demonstrate the performance of our
approach.

\subsection{Results and Discussion}
\subsubsection{Motion planning through dynamical system relaxation}
We focus on finding trajectories that are only feasible when dynamics and
contact forces are considered. Our motion planning approach describes contact
events in a \gls{mpcc} problem. Since the problem is non-convex, our
hierarchical optimization tends to guide the exploration away from infeasible
regions through dynamical system relaxation, i.e. centroidal to full dynamics.
This dynamical relaxation helps to reduce the computation time and
cost value.%

Our experiments suggest that dynamical system relaxation is key for finding
successful motion plans. Table \ref{tab:results} shows the time and cost
reduction of our approach compared with a single full trajectory optimization.
We can see that the hierarchical optimization approach tends to have better
performance in complex tasks. Nevertheless, in general, the central tendency
(median) of the computation time reduction is decreased, while, on average, we
improve the quality of the solution (cost reduction). We define a set of
8 different goal states (i.e. trunk height) for computing the time and cost
reduction of our approach. For computing the time and cost reduction, we compare
the time and cost of a single full trajectory optimization trial against our hierachical optimization approach in the set of goals defined.

\begin{table}[!t]
\caption{Time and cost reduction over 8 trials compared to a single full
trajectory optimization.}
 \label{tab:results} 
\begin{center}
\begin{tabular}{@{} lrrrrr @{}}    
	\toprule
	&\multicolumn{2}{c}{Time reduction [\%]} && \multicolumn{2}{c}{Cost reduction [\%]}\\
	\cmidrule{2-3}\cmidrule{5-6}
	\emph{Task}		&Md.		&Av.	&&Md.	&Av.\\
	\midrule
	Jumping			&10.36		&0.0	&&0.0	&1.44\\ 
	Step Jumping	&48.29		&30.46	&&0.0	&12.91\\
	\bottomrule
\end{tabular}\vspace{-0.3cm}
\end{center}
\end{table}

\subsubsection{Reaching goals that are kinematically not feasible}
The jumping task demonstrates the ability of exploring the dynamical
capabilities of the robot in order to reach goals that are not kinematically
possible. In this particular case, we desire to reach with the trunk, a height
of \unit[0.85]{cm} w.r.t. the ground (or \unit[0.27]{cm} w.r.t. the initial
position), which is kinematically not feasible. Thus, our hierarchical motion
planner explores different mode sequences in order to plan a dynamically
feasible motion. In Fig. \ref{fig:jumping_with_posture}, the robot plans a
countermovement jump (around \unit[7]{cm}) without being predefined. In
countermovement jumps, a preliminary downward movement is executed which
increases the jump height because the robot is carried by its own inertia. Then,
a fast movement of the foot is planned considering a desired task behavior, e.g.
joint position in the apex point.

\subsubsection{Discovering of new contacts}
For the success of some tasks, it is crucial to exploit the environmental
conditions, e.g. reaching and keeping a desired trunk position that is
kinematically not reachable. So, imagine that we want to keep a desired trunk
position but due to gravitational forces this is not possible with just a vertical
jump, for instance we need to climb onto an obstacle to accomplish this. 
With the hierarchical trajectory optimization, we can plan such kind of
maneuvers. In fact, Fig. \ref{fig:step_jumping} shows that our motion planner
solves these tasks by defining a foolhold on top of an available step. 
Note that a pre-defined footstep sequence is not required to arrive to such a
solution.%

%% file: src/conclusion.tex
\section{Conclusion}\label{sec:conclusion}
In this paper we presented a hierarchical trajectory optimization approach for
planning dynamic movements through unscheduled contact sequence. First, the
hierarchical trajectory optimization finds a feasible \gls{com} motion according
to the centroidal dynamics of the robot. Then, a second phase of optimization
considers the full-dynamics of the robot. In both phases we use complementarity
constraints for contact modeling. We demonstrated that, with our approach, the
robot can plan a wide range of movements that consider the full-dynamics and
joint effort limits of the robot. We believe that these considerations are
crucial for highly-dynamic locomotion tasks, i.e. \textit{step-jumping} tasks
that are cannot be accomplished in a kinematic fashion. We showed how the
hierarchical trajectory optimization improves the solutions and significantly
reduces the computation time, compared with the full dynamic optimization.
Experimental trials with a robotic leg performing highly-dynamic and challenging
tasks demonstrate the capability of our planning approach.

In general, trajectory optimization produces motion plans for a fixed
sequence of points, $N$ knot-points. Here, we arrive at a continuous motion plan
by applying a polynomial interpolation. Nevertheless, polynomial interpolations
cannot predict accurately changes in the contact forces, which generally happen
in less than \unit[10]{ms}. We are currently working on generating
motion plans given a library of synthesized motions, improving the quality of
the solutions, and permitting on-line computation and execution.

\begin{figure}[t]
	\centering
	\includegraphics[height=0.365\textheight]{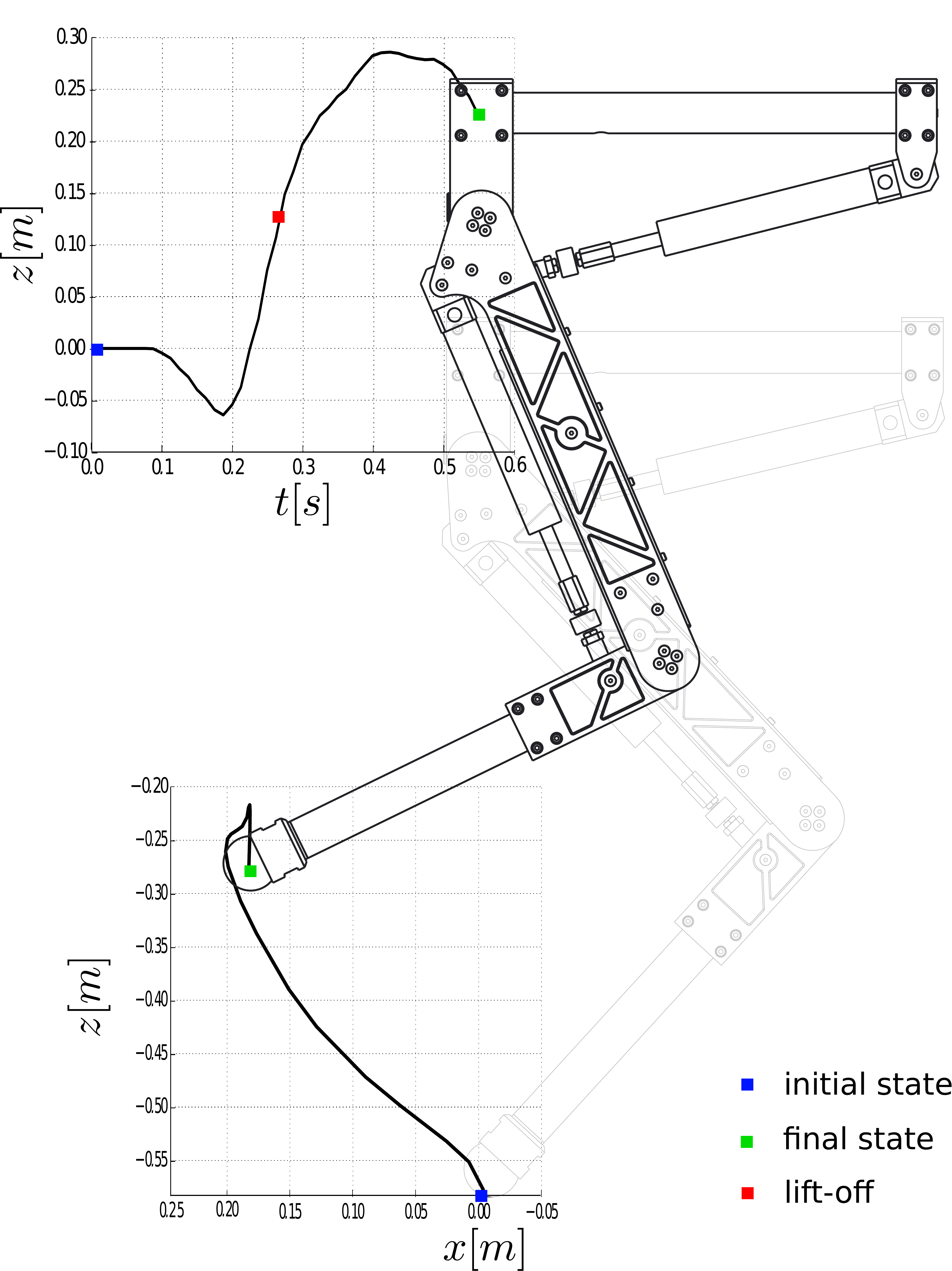}
	\caption{Optimized CoM and foot trajectory for a jumping task that shows a
	dynamic movement through different phases: thrust and flight. We can see
	that the hierarchical optimization maximizes the jump energy by planning a
	countermovement jump, i.e. a preliminary downward movement.}\vspace{-0.2cm}
 	\label{fig:jumping_with_posture}
\end{figure} 

\begin{figure}[t]
	\centering
	\includegraphics[height=0.365\textheight]{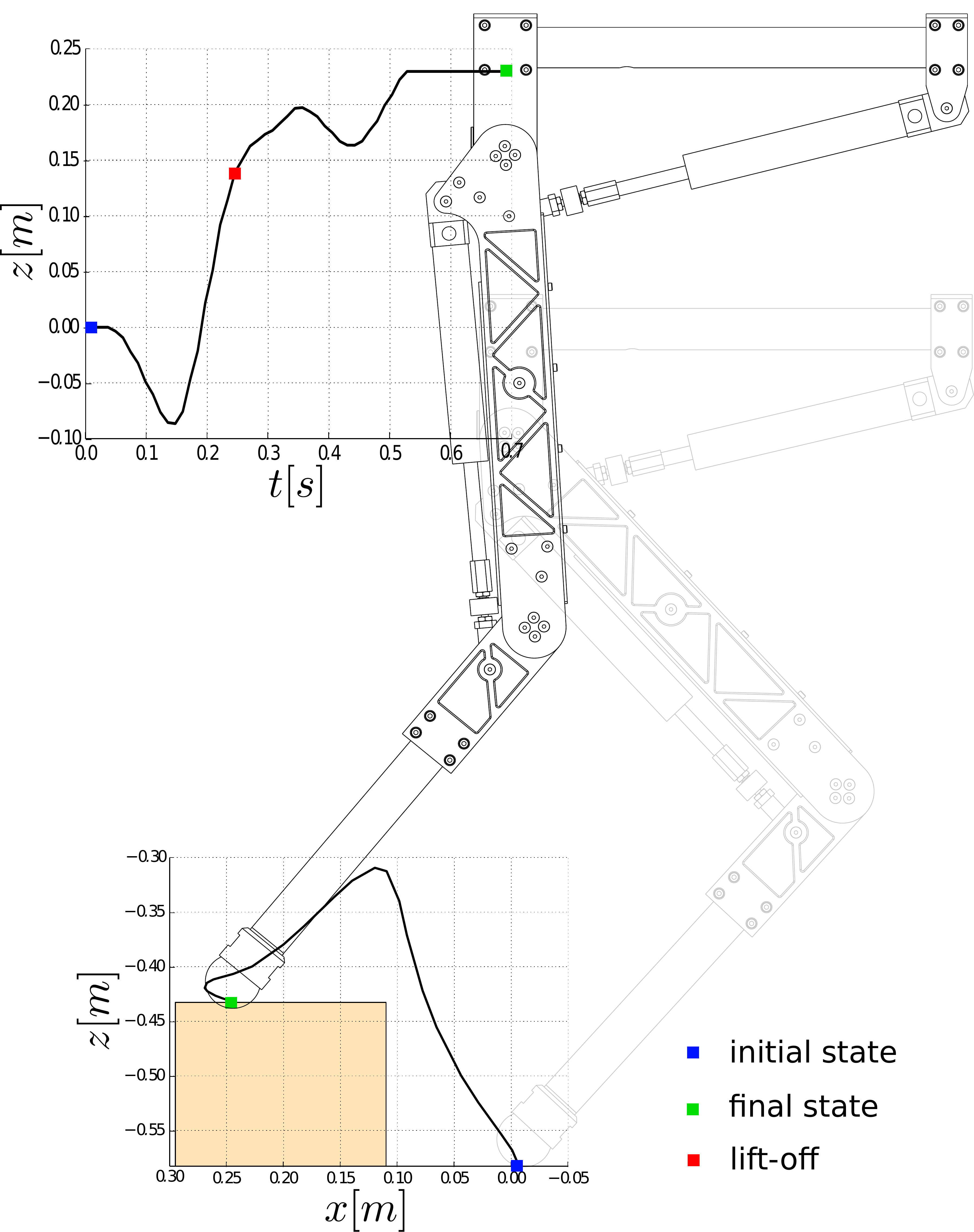}
	\caption{Optimized CoM and contact sequence for reaching and keeping a
	desired trunk position (big step jumping), which is not reachable with just a
	vertical motion. The hierarchical trajectory optimization finds a
	transitory foothold in order to keep the desired trunk position. 
	After landing
	in the planned foothold, the trunk moves up until the desired
	goal.}\vspace{-0.2cm}
 	\label{fig:step_jumping}
\end{figure}\vspace{0.15em}